%% file: paper.tex
\documentclass{article}

\usepackage{microtype}
\usepackage{graphicx}
\usepackage{subfigure}
\usepackage{booktabs}
\usepackage{hyperref}
\usepackage{amsmath}
\usepackage{multirow}
\usepackage{natbib}

\usepackage[accepted]{icml2020}

\icmltitlerunning{Deep Transformer Models for Time Series Forecasting}

\begin{document}
\twocolumn[
\icmltitle{Deep Transformer Models for Time Series Forecasting:\\The Influenza Prevalence Case}

\begin{icmlauthorlist}
\icmlauthor{Neo Wu}{goo}
\icmlauthor{Bradley Green}{goo}
\icmlauthor{Xue Ben}{goo}
\icmlauthor{Shawn O'Banion}{goo}
\end{icmlauthorlist}

\icmlaffiliation{goo}{Google, LLC, 651 N 34th St., Seattle, WA 98103 USA}

\icmlcorrespondingauthor{Neo Wu}{neowu@google.com}
\icmlcorrespondingauthor{Bradley Green}{brg@google.com}
\icmlcorrespondingauthor{Xue Ben}{sherryben@google.com}
\icmlcorrespondingauthor{Shawn O'Banion}{obanion@google.com}

\icmlkeywords{transformer, time series, forecasting, influenza}

\vskip 0.3in
]

\printAffiliationsAndNotice{}

\begin{abstract}
In this paper, we present a new approach to time series forecasting.
Time series data are prevalent in many scientific and engineering disciplines.
Time series forecasting is a crucial task in modeling time series data,
and is an important area of machine learning.
In this work we developed a novel method that employs Transformer-based machine learning models to forecast time series data.
This approach works by leveraging self-attention mechanisms to learn complex patterns and dynamics from time series data.
Moreover, it is a generic framework and can be applied to univariate and multivariate time series data,
as well as time series embeddings.
Using influenza-like illness (ILI) forecasting as a case study,
we show that the forecasting results produced by our approach are favorably comparable to
the state-of-the-art.
\end{abstract}

\section{Introduction}
\input{intro.tex}

\begin{figure*}
  \includegraphics[width=\textwidth]{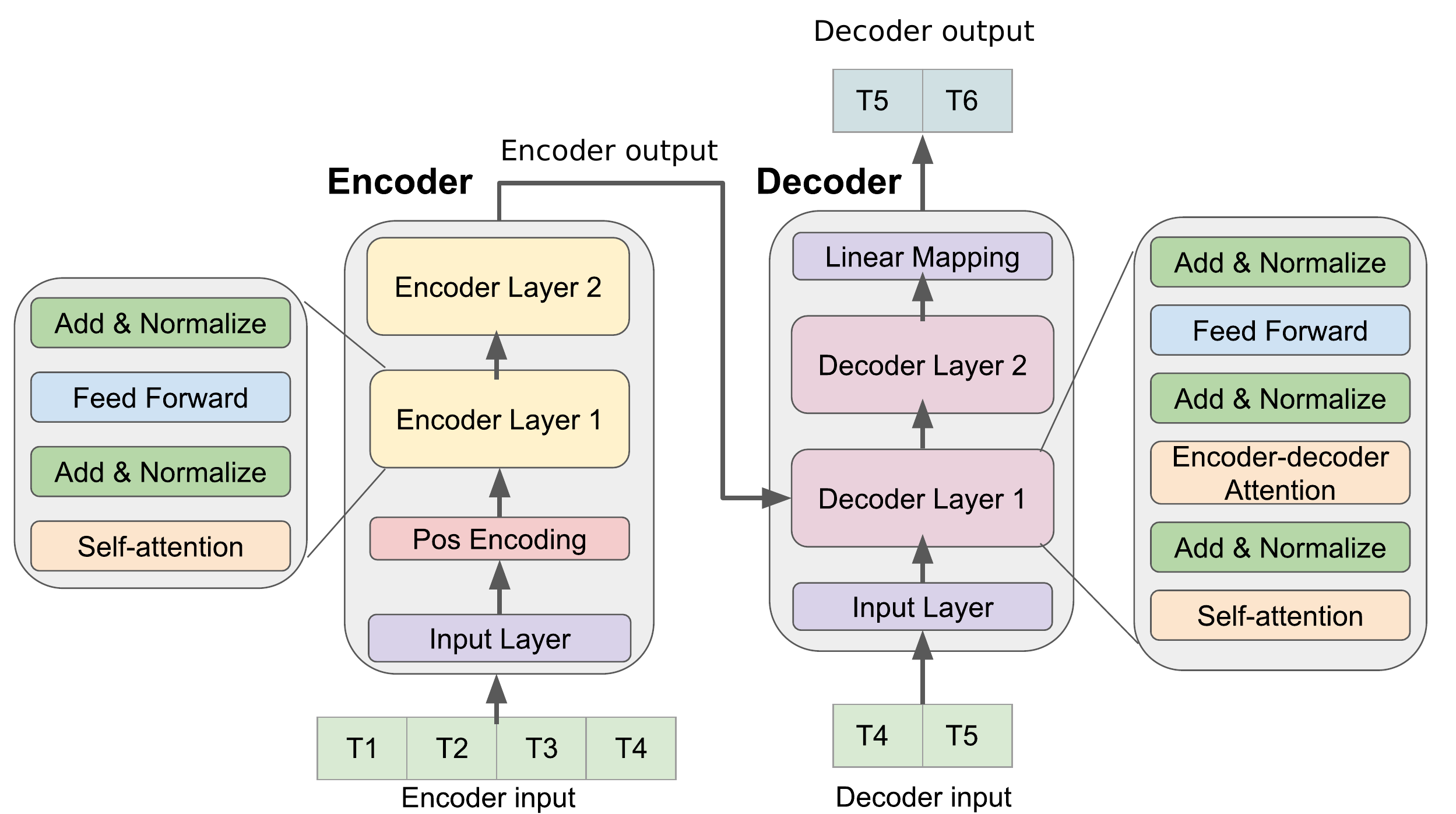}
  \caption{Architecture of Transformer-based forecasting model.}
\end{figure*}

\section{Related Work}
\input{related_work.tex}

\section{Background}
\input{background.tex}

\section{Model}
\input{model.tex}

\section{Experiment}
\input{experiment.tex}
\section{Conclusions}
In this work, we presented a Transformer-based approach to forecasting time series data.
Compared to other sequence-aligned deep learning methods, our approach leverages self-attention mechanisms to model sequence data,
and therefore it can learn complex dependencies of various lengths from time series data.

Moreover, this Transformer-based approach is a generic framework for modeling various non-linear dynamical systems.
As manifested in the ILI case,
this approach can model observed time series data as well as phase space of state variables through time delay embeddings.
It is also extensible and can be adpated to model both univariate and multivariate time series data
with minimum modifications to model implementations.

Finally, although the current case study focuses on time series data, we hypothesize that our approach can be
further extended to model spatio-temporal data indexed by both time and location coordinates.
Self-attention mechanisms can be generalized to learn relations between two arbitrary points in spatio-temporal space.
This is a direction we plan to explore in the future.
\bibliography{paper}
\bibliographystyle{icml2019}
\end{document}

%% file: intro.tex
Seasonal influenza epidemics create huge health and economic burdens, causing 291,000 - 646,000 deaths worldwide \cite{Iuliano2018}.
In the United States, the Centers for Disease Control and Prevention (CDC) publishes weekly ILI reports based on its surveillance network.
Despite its significance in monitoring disease prevalence, typically there is at least a one-week delay of ILI reports due to data collection and aggregation.
Therefore, forecasting ILI activity is critical for real-time disease monitoring,
and for public health agencies to allocate resources to plan and prepare for potential pandemics.

A variety of methods have been developed to forecast these ILI time series data.
These approaches range from mechanistic approaches to statistical and machine learning methods.
Mechanistic modeling is based on the understanding of underlying disease infection dynamics.
For example, compartmental methods such as SIR are popular approaches to simulating disease spreading dynamics.

Statistical and machine learning methods leverage the ground truth data to learn the trends and patterns.
One popular family of methods includes auto-regression (AR), autoregressive moving average (ARMA), and autoregressive integrated moving average (ARIMA).
Additionally, deep learning approaches based on convolutional and recurrent neural networks have been developed to model ILI data.
These sequence-aligned models are natural choices for modeling time series data.
However, due to ``gradient vanishing and exploding" problems in RNNs and the limits of convolutional filters,
these methods have limitations in modeling long-term and complex relations in the sequence data.

In this work, we developed a novel time series forecasting approach based on Transformer architecture \cite{Vaswani17}.
Unlike sequence-aligned models, Transformer does not process data in an ordered sequence manner.
Instead, it processes entire sequence of data and uses self-attention mechanisms to learn dependencies in the sequence.
Therefore, Transformer-based models have the potential to model complex dynamics of time series data that are challenging for sequence models.
In this work we use ILI forecasting as a case study to show that a Transformer-based model can be successfully applied to the task of times series forecasting
and that it outperforms many existing forecasting techniques. Specifically, our contributions are the following:
\begin{itemize}
  \item We developed a general Transformer-based model for time series forecasting.
  \item We showed that our approach is complementary to state space models. It can model observed data. Using embeddings as a proxy, our approach can also model state variables and phase space of the systems.
  \item Using ILI forecasting as a case study, we demonstrated that our Transformer-based model is able to accurately forecast ILI prevalence using a variety of features.
  \item We showed that in the ILI case our Transformer-based model achieves state-of-the-art forecasting results.
\end{itemize}

%% file: related_work.tex
Several studies have used Internet data such as Google Trends\cite{Ginsberg2009}, Twitter \cite{Paul2014} and Wikipedia \cite{McIver2014} to forecast ILI ratios.
Google Flu Trends (GFT) employs a linear model that uses Google search volumes of predefined terms to estimate current ILI ratios (``nowcasting").
While initially regarded as a great success, GFT suffered from over-estimation of peak ILI magnitude in subsequent years \cite{Olson2013, Lazer2014}.

Other studies combined GFT with new modeling techniques and additional signals.
Lazer \textit{et~al} \yrcite{Lazer2014} suggested an autoregression-based (AR) approach to extend GFT.
Santillana \textit{et~al} \yrcite{Santillana2014} improved GFT by developing a model that automatically selects queries and updates models for ILI forecasting.
Araz \textit{et~al} \yrcite{Araz2014} built linear regression models using GFT data with extra signals.
Yang \textit{et~al} \yrcite{Yang2015} developed an autoregression-based model with Google search data (``ARGO") to estimate influenza epidemics.
ARGO outperforms previous Google-search-based models.
More recently, a new ensemble model (``ARGONet") built on top of ARGO was developed \cite{Lu2019}.
This new approach leverages spatial information to improve the model
and achieved state-of-the-art results for ILI forecasting.

Deep learning techniques have also been used for ILI forecasting.
Liu \textit{et~al} \yrcite{Liu:Han:2018} trained an LSTM-based model to predict influenza prevalence using Google Trends, climate, air pollution and virological surviellence data.
Venna \textit{et~al} \yrcite{Venna2019} developed an LSTM-based multi-stage model to incorporate climate and spatio-temporal adjustment factors for influenza forecasting.
Attention-based technqiues are also applied for ILI forecasting.
Zhu \textit{et~al} \yrcite{Zhu2019} developed multi-channel LSTM neural networks to learn from different types of inputs.
Their model uses an attention layer to associate model output with the input sequence to further improve forecast accuracy.
Kondo \textit{et~al} \yrcite{Kondo2019} adapted a sequence-to-sequence (``Seq2Seq") model with a similar attention mechanism to predict influenza prevalence
and showed that their approach outperformed ARIMA and LSTM-based models.

%% file: background.tex
\subsection{Influenza and ILI}
Influenza is a common infectious disease caused by infection of influenza virus.
Influenza affects up to 35 million people each year and creates huge
health and economic burdens \cite{Iuliano2018, Putri2018}. In the United States, the
Centers for Disease Control and Prevention (CDC) coordinates a
reporting network over a large geographic area.
Participating health providers within the network report statistics of patients
showing influenza-like illness (ILI) symptoms.
ILI symptoms are usually defined as \emph{fever and cough and/or sore throat}.
The ILI ratio is computed as the ratio of
the number of patients seen with ILI and the total number of patient visits that calendar week.
The CDC published ILI ratios for the USA and all but one individual state (Florida).
Additionally, state-level ILI ratios are normalized by state populations.

\subsection{State Space Models}
State space modeling (SSM) is widely applied to dynamical systems.
The evolution of dynamical systems is controlled by non-observable state variables.
The system exhibits observable variables which are determined by state variables.
SSM has been applied to study complex systems in biology and finance.

State space models model both state and observable variables.
For example, a generalized linear state space model can be expressed in the following equations:
\begin{align}
x_t &= Z_t\alpha_t + \epsilon_t  \label{obs:eqn} \\
\alpha_{t+1} &= T_t\alpha_t + R_t\eta_t,  t = 1, ..., n, \label{state:eqn}
\end{align}
where $x_t$ and $\alpha_t$ are time-indexed observation vectors and state vectors, respectively.
Equation ~\ref{obs:eqn}, called observation equation, is a regression-like equation.
It models the relationship of observable $x_t$ and the underlying state variable $\alpha_t$.
Equation ~\ref{state:eqn} is the state equation, and has autoregressive nature.
It governs how the state variables evolve over time.
$\epsilon_t$ and $\eta_t$ are innovation components and are usually modeled as Gaussian processes.

In this section, we briefly mention a few commonly used SSM models in ILI forecasting.

\paragraph*{Compartmental Models}
Compartmentmental models are a specific form of SSMs and have been widely used to study infectious diseases.
In a compartmental model, a population is divided into different groups (``compartments").
Each group is modeled by a time-dependent state variable.
An prominent example of compartmental model is ``Suscepted-Infected-Recovered" (SIR) model,
where the system is governed by three state variables ($S(t)$, $I(t)$, $R(t)$) through the following ordinary differential equations:
\begin{align*}
  \frac{dS}{dt} &= -\frac{\beta IS}{N}\\
  \frac{dI}{dt} &= \frac{\beta IS}{N} - \gamma I\\
  \frac{dR}{dt} &= -\gamma I
\end{align*}
In this treatment, ILI time series is an observable variable of the system: $\mathrm{ILI}(t) = I(t) / (I(t) + S(t) + R(t))$.

Although originally developed to model infectious diseases, compartmental models have been applied to other disciplines such as ecology and economics.
While compartmental models are useful, they require prior knowledge on the parameters of the differential equations
and they lack flexibility of updating parameters upon new observations.

\paragraph*{ARIMA}
Box-Jenkins ARIMA (Auto-Regressive Integrated Moving Average) is another popular approach to modeling dynamical systems.
ARIMA models the observed variable $x_t$ and assumes $x_t$ can be decomposed into trend, seasonal and irregular components.
Instead of modeling these components separately, Box and Jenkins had the idea of differencing the time series $x_t$
in order to eliminate trend and seasonality.
The resulting series is treated as stationary time series data and is modeled using combination of its lagged time series values (``AR") and moving average of lagged forecast errors (``MA").
An ARIMA model is typically specified by a tuple $(p, d, q)$, where $p$ and $q$ define the orders of AR and MA, and $d$ specifies the order of differencing operation.

ARIMA can be written in SSM form, and common SSM techniques such as filtering and smoothing can be applied to ARIMA as well.
Nevertheless, ARIMA is a kind of ``blackbox" approach where the model purely depends on the observed data
and has no analysis of the states of the underlying systems \cite{DurbinKoopman2012}.

\paragraph* {Time Delay Embedding}
For a scalar time series data $x_t$, its time delay embedding (TDE) is formed by embedding each scalar value $x_t$ into a $d$-dimensional time-delay space:
\begin{equation*}
\mathrm{TDE_{d, \tau}}(x_t) = (x_t, x_{t-\tau}, ..., x_{t-(d-1)\tau}) % TODO(neowu) check if whether this the right definition?
\end{equation*}
For any non-linear dynamical systems, the delay-embedding theorem (Takens' theorem) \cite{Takens1981} states that
there exists a certain $(d, \tau)$-time delay embedding
such that the evolution of the original state variables (``phase space") can be recovered in the delay coordinates of the observed variables.
In the case of ILI forecasting, Takens' theorem suggests that $\mathrm{TDE_{d, \tau}}$ of ILI ratios (``observed variable")
can approximate the underlying dynamical systems governed by biological and physical mechanisms.

TDEs were first explored for time series forecasting in the seminal work by Sugihara and May \yrcite{Sugihara1990}.
They showed that TDEs can be used to make short-range predictions based on the qualitative assessment of a system's dynamics without any knowledge on the underlying mechanisms.
They developed two TDE-based models to predict chickenpox and measles prevalence, and compared them with AR-based approaches.
Their analysis suggests that TDE-based model performs equally well for chickenpox case prediction and outperforms AR for measles case prediction.

In the SSM framework, the time delay embedding is a powerful tool to bridge state variables and observed data by making it
possible to learn geometrical and topological information of underlying dynamical systems without an understanding of the state variables and phase space of the systems.
Depite the amazing property of TDEs, to the best of our knowledge, TDEs haven't been studied extensively for machine learning models.

\subsection{Sequence Models}
Many real-world machine learning tasks deal with different types of sequential data ranging from natural language text, audio, and video, to DNA sequences and time series data.
Sequence models are specifically designed to model such data.
In this section, we briefly review a few different types of common sequence models.

\paragraph*{Recurrent Neural Networks}
Unlike traditional feed-forward networks, RNN is recurrent in nature - it performs the same function to each input $x_t$,
and the output $y_t$ depends on both the input $x_t$ and the previous state  $h_{t-1}$.

\begin{figure}
\centering
\label{fig:RNN}
\includegraphics[width=8cm,height=5cm,keepaspectratio]{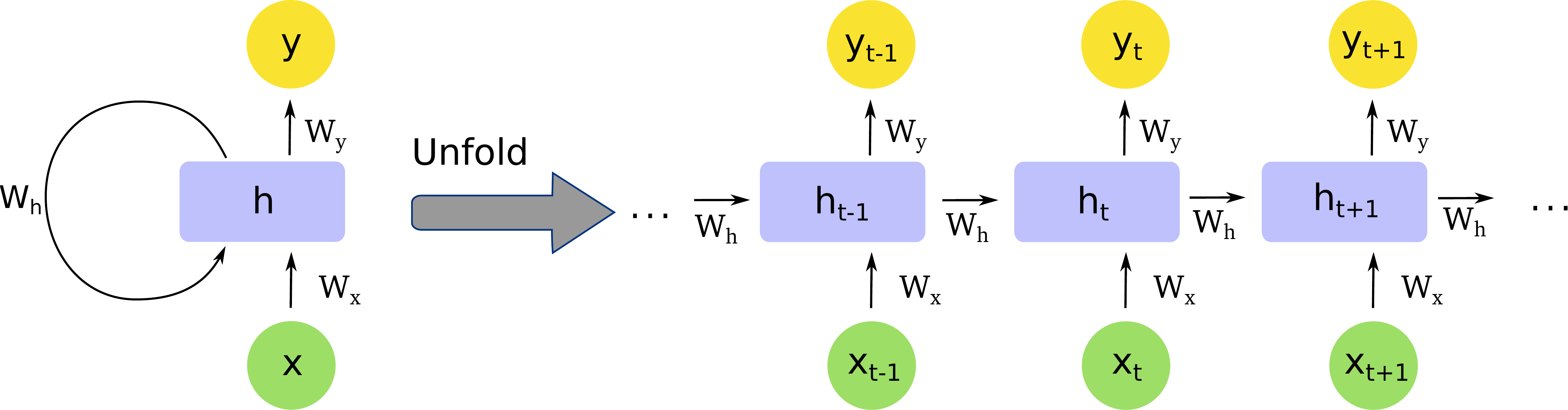}
\caption{Folded and unfolded representations of recurrent neural network.}
\end{figure}

The simple RNN illustrated in Figure ~\ref{fig:RNN} can be expressed as follows:
\begin{align*}
  h_t &= \sigma(W_x x_t + W_h h_{t-1} + b_h) \\
  y_t &= \sigma(W_y h_t + b_y)
\end{align*}
Where $x_t$ is the input vector, $h_t$ is the hidden state vector, and $y_t$ is the output vector.
$W$'s and $b$'s are learned parameters, and $\sigma$ is the activation function.

\paragraph*{LSTM}
While RNN has internal memory to process sequence data, it suffers from gradient vanishing and exploding problems when processing long sequences.
Long Short-Term Memory (LSTM) networks were specifically developed to address this limitation \cite{hochreiter1997}.
LSTM employs three gates, including an input gate, forget gate and output gate, to modulate the information flow across the cells and prevent gradient vanishing and explosion.

\begin{figure}
\centering
\includegraphics[width=8cm,height=5cm,keepaspectratio]{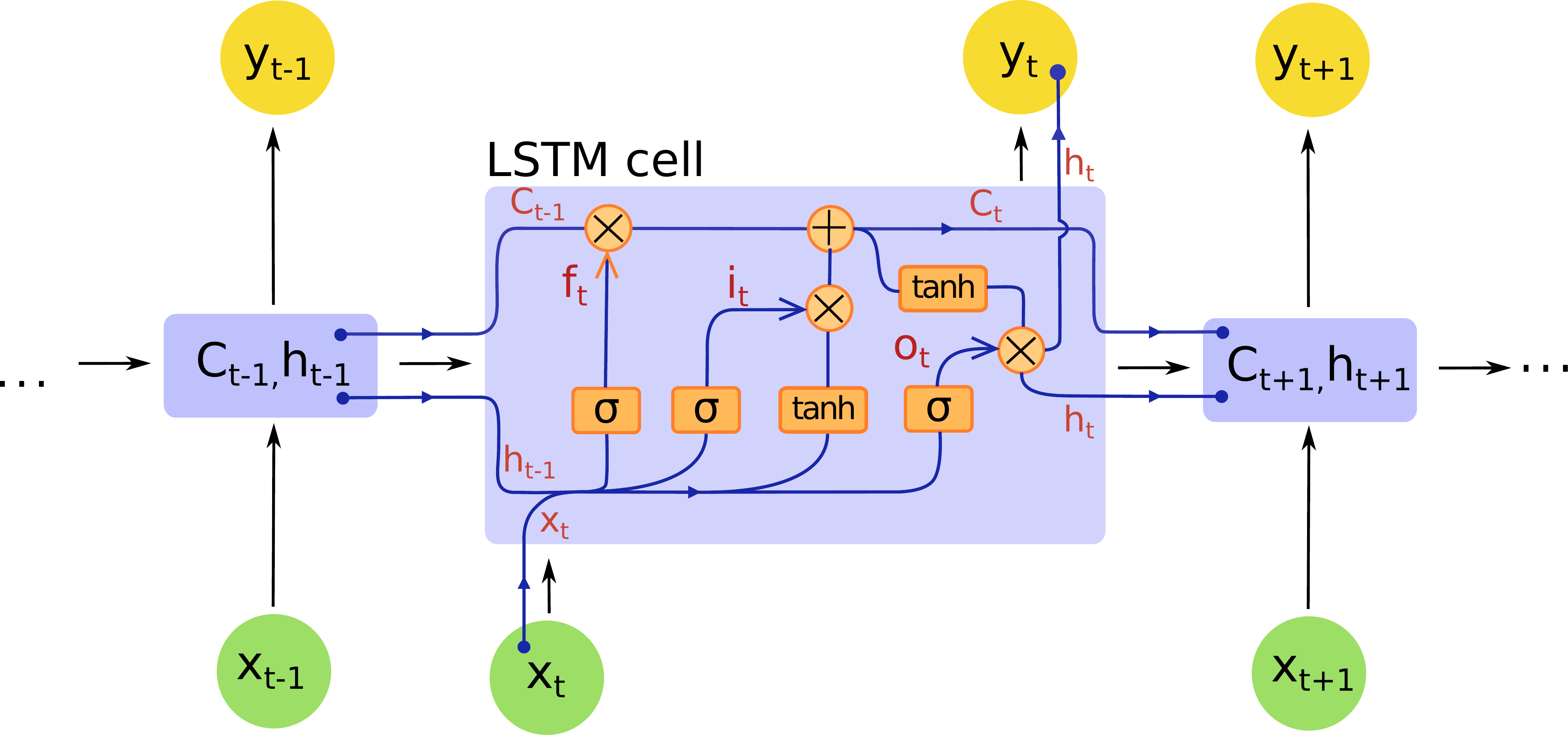}
\caption{Long Short-Term Memory network and LSTM unit.}
\end{figure}
\begin{align*}
  f_t &= \sigma(W_f [h_{t-1}, x_t] + b_f) \\
  i_t &= \sigma(W_i [h_{t-1}, x_t] + b_i) \\
  \tilde{C_t} &= tanh(W_C [h_{t-1}, x_t] + b_C) \\
  C_t &= f_t * C_{t-1} + i_t * \tilde{C_t} \\
  y_t &= \sigma(W_y [h_{t-1}, x_t] + b_y) \\
  h_t &= y_t * tanh(C_t)
\end{align*}

\begin{figure}
\centering
\label{fig:seq2seq}
\includegraphics[width=8cm,height=5cm,keepaspectratio]{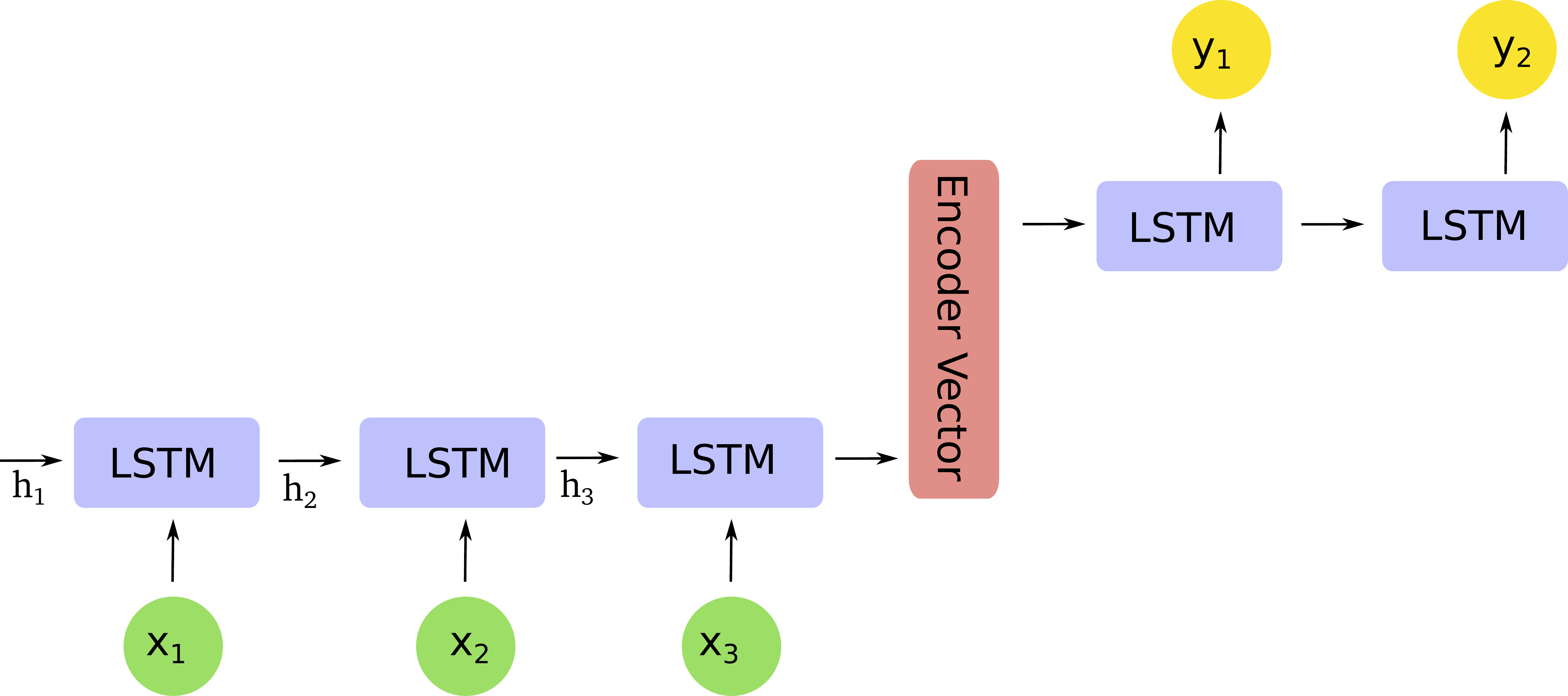}
\caption{Sequence-to-sequence (Seq2Seq) architecture to model sequence input and output.}
\end{figure}

\paragraph*{Seq2Seq}
Sequence-to-sequence (Seq2Seq) architecture is developed for machine learning tasks where both input and output are sequences.
A Seq2Seq model is comprised of three components including an encoder, an intermediate vector, and a decoder.
Encoder is a stack of LSTM or other recurrent units. Each unit accepts a single element from the input sequence.
The final hidden state of the encoder is called the encoder vector or context vector, which encodes all of the information from the input data.
The decoder is also made of a stack of recurrent units and takes the encoder vector as its first hidden state.
Each recurrent unit computes its own hidden state and produces an output element.
Figure ~\ref{fig:seq2seq} illustrates the Seq2Seq architecture.

Seq2Seq has been widely applied in language translation tasks.
However, its performance degrades with long sentences because it cannot adequately encode a long sequence into the intermediate vector (even with LSTM cells).
Therefore, long-term dependencies tend to be dropped in the encoder vector.

%% file: model.tex
\subsection{Problem Description}
We formuate ILI forecasting as a supervised machine learning task.
Given a time series containing $N$ weekly data points ${x_{t-N+1}, ..., x_{t-1}, x_t}$, for M-step ahead prediction,
the input $X$ of the supervised ML model is  ${x_{t-N+1}, ..., x_{t-M}}$, and the output $Y$ is ${x_{t-M+1}, x_{t-M+2}, ..., x_t}$.
Each data point $x_t$ can be a scalar or a vector containing multiple features.

\subsection{Data}
We utilized country- and state-level historical ILI data from 2010 to 2018 from the CDC \cite {CDC_website}.

\begin{figure}
  \includegraphics[width=8cm,height=8cm,keepaspectratio]{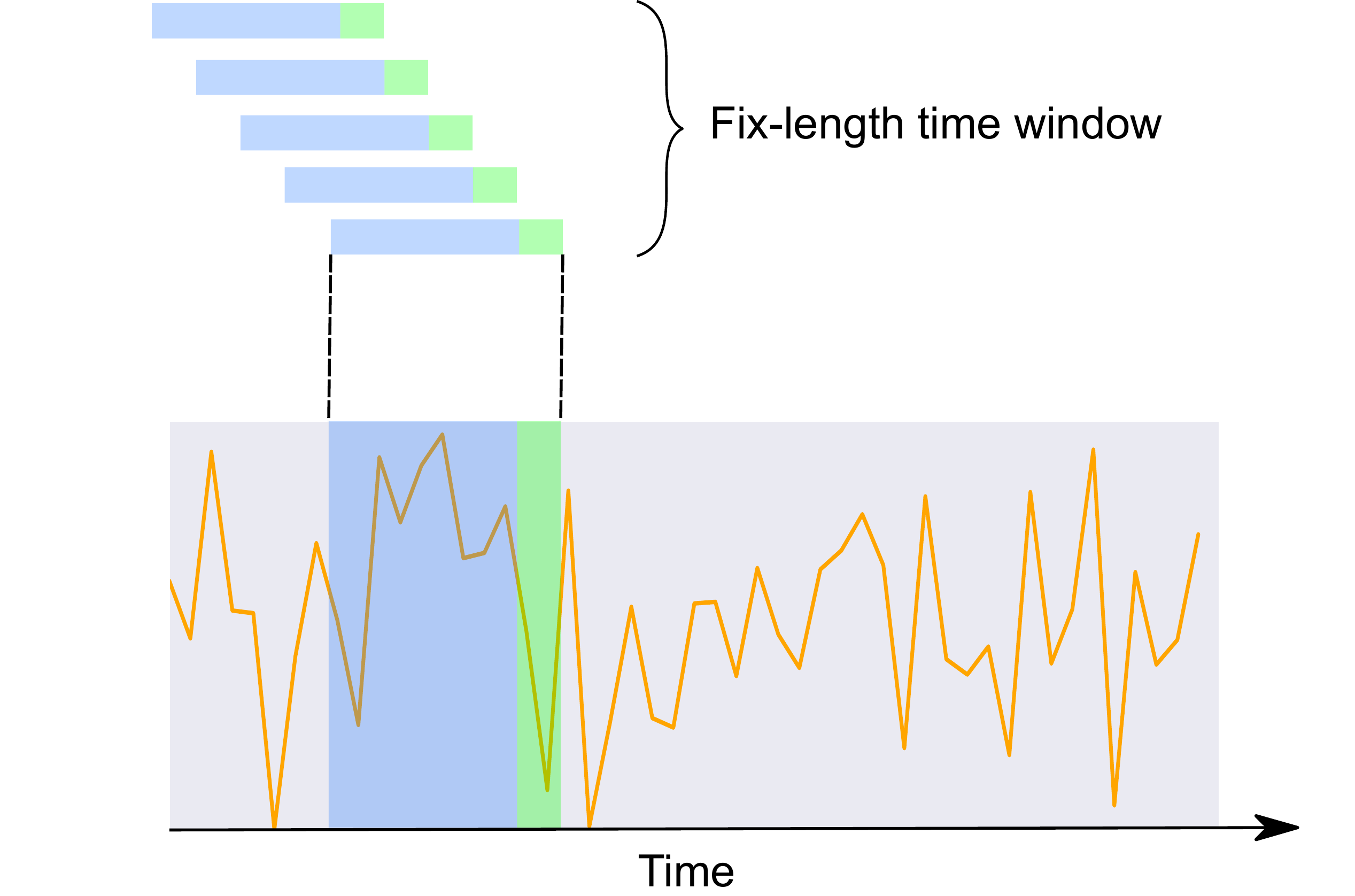}
  \caption{Using sliding window to construct supervised learning examples from time series data.}
  \label{fig:sliding-window}
\end{figure}

To produce a labeled dataset, we used a fixed-length sliding time window approach (Figure ~\ref{fig:sliding-window})
to construct ${X, Y}$ pairs for model training and evaluation.
Before applying the sliding window to get features and labels, we perform min-max scaling on all the data with the maximum and minimum values of training dataset.
We then run a sliding window on the scaled training set to get training samples with features and labels, which are the previous N and next M observations respectively.
Test samples are also constructed in the same manner for model evaluation. The train and test split ratio is 2:1.
Training data from different states are concatenated to form the training set for the global model.

\subsection{Transformer Model}
\subsubsection {Model Architecture}
Our Transformer-based ILI forecasting model follows
the original Transformer architecture \cite{Vaswani17} consisting of encoder and decoder layers.

\paragraph*{Encoder}
The encoder is composed of an input layer, a positional encoding layer, and a stack of four identical encoder layers.
The input layer maps the input time series data to a vector of dimension $d_{model}$ through a fully-connected network.
This step is essential for the model to employ a multi-head attention mechanism.
Positional encoding with sine and cosine functions is used to encode sequential information in the time series data
by element-wise addition of the input vector with a positional encoding vector.
The resulting vector is fed into four encoder layers. Each encoder layer consists of two sub-layers:
a self-attention sub-layer and a fully-connected feed-forward sub-layer.
Each sub-layer is followed by a normalization layer.
The encoder produces a $d_{model}$-dimensional vector to feed to the decoder.

\paragraph*{Decoder}
We employ a decoder design that is similar to the original Transformer architecture \cite{Vaswani17}.
The decoder is also composed of the input layer, four identical decoder layers, and an output layer.
The decoder input begins with the last data point of the encoder input.
The input layer maps the decoder input to a $d_{model}$-dimensional vector.
In addition to the two sub-layers in each encoder layer, the decoder inserts a third sub-layer to
apply self-attention mechanisms over the encoder output. Finally, there is an output layer that maps
the output of last decoder layer to the target time sequence.
We employ look-ahead masking and one-position offset between the decoder input and target output in the decoder
to ensure that prediction of a time series data point will only depend on previous data points.

\subsubsection{Training}
\paragraph*{Training Data and Batching}
In a typical training setup, we train the model to predict 4 future weekly ILI ratios from
10 trailing weekly datapoints. That is,
given the encoder input $(x_1, x_2, ..., x_{10})$ and the decoder input $(x_{10}, ..., x_{13})$, the decoder aims to output $(x_{11}, ..., x_{14})$.
A look-ahead mask is applied to ensure that attention will only be applied to datapoints prior to target data by the model.
That is, when predicting target $(x_{11}, x_{12})$, the mask ensures attention weights are only on $(x_{10}, x_{11})$ so the decoder doesn't leak information about $x_{12}$ and $x_{13}$ from the decoder input.
A minibatch of size 64 is used for training.

\paragraph*{Optimizer}
We used the Adam optimizer \cite{Kingma2015} with $\beta_1 = 0.9$, $\beta_2 = 0.98$ and $\epsilon = 10^{-9}$.
A custom learning rate with following schedule is used:
\begin{align*}
lrate = &d_{model}^{0.5} * \min(step\_num^{0.5},\\
        &step\_num * warmup\_steps^{-1.5})
\end{align*}
Where $warmup\_steps = 5000$.

\paragraph*{Regularization}
We apply dropout techniques for each of the three types of sub-layers in the encoder and decoder:
the self-attention sub-layer, the feed-forward sub-layer, and the normalization sub-layer.
A dropout rate of 0.2 is used for each sub-layer.

\subsubsection{Evaluation}
In evaluation, labeled test data are constructed using a fix-length sliding window as well.
One-step ahead prediction is performed by the trained Transformer model.
We computed Pearson correlation coefficient and root-mean-square errors (RMSE)
between the actual data ${y_i}$ and the predicted value ${\hat{y_i}}$.

\subsection{ARIMA, LSTM and Seq2Seq Models}
This section describes other models we developed to benchmark Transformer-based model.
\paragraph*{ARIMA}
A univariate ARIMA model is used as a baseline. It treats the time dependent
ILI ratios as a univariate time series that follows a fixed dynamic. Each
week's ILI ratio is dependent on previous $p$ weeks' observations and previous $q$ weeks'
estimation errors. We selected the order of ARIMA model using AIC and BIC to balance
model complexity and generalization. We used $\mathrm{ARIMA}(3,0,3)$ and a constant trend to
keep the model parsimonious. The model is formulated in the State Space
modeling framework and trained with the first two-thirds of the dataset. The fitted
parameters are then used on the full time series to filter hidden states and make four-step
ahead predictions.

\paragraph*{LSTM}
The LSTM model has a stack of two LSTM layers and a final dense layer to predict
the multiple step ILI ratios directly. The LSTM layers encode sequential information
from input through the recurrent network. The dense connected layer takes final
output from the second LSTM layer and outputs a vector of size 4, which is equal to the
number of steps ahead predictions. The two LSTM layers are 32 and
16 units respectively. A dropout rate 0.2 is applied to LSTM layers for regularization.
Huber loss, Adam optimizer, and a learning rate of 0.02 are used for training.

\paragraph*{Seq2Seq}
The tested Seq2Seq model has an encoder-decoder architecture, where the encoder
is composed of a fully connected dense layer and a GRU layer to learn from the
sequential input and to return a sequence of encoded outputs and a final hidden state.
The decoder is of the same structure as input.
The dense layer is of 16 units and the GRU layer is of 32 units.
An attention mechanism is also adopted in this Seq2Seq model. Specifically, Bahdanau
attention \cite{bahdanau14ICLR} is applied on the sequence of encoder
outputs at each decoding step to make next step prediction. Teacher forcing
\cite{williams89NC} is utilized in decoder for faster convergence and to address
instability. During training, the true ILI rate is used at current time step as
input for the next time step, instead of using output computed from the decoder
unit. A dropout rate 0.2 is applied in all recurrent layers.
We used Huber loss, Adam optimizer, and a learning rate of 0.02 for training.

%% file: experiment.tex
\subsection{One-step-ahead Forecasting Using ILI Data Alone}
\begin{figure*}
\includegraphics[width=\textwidth]{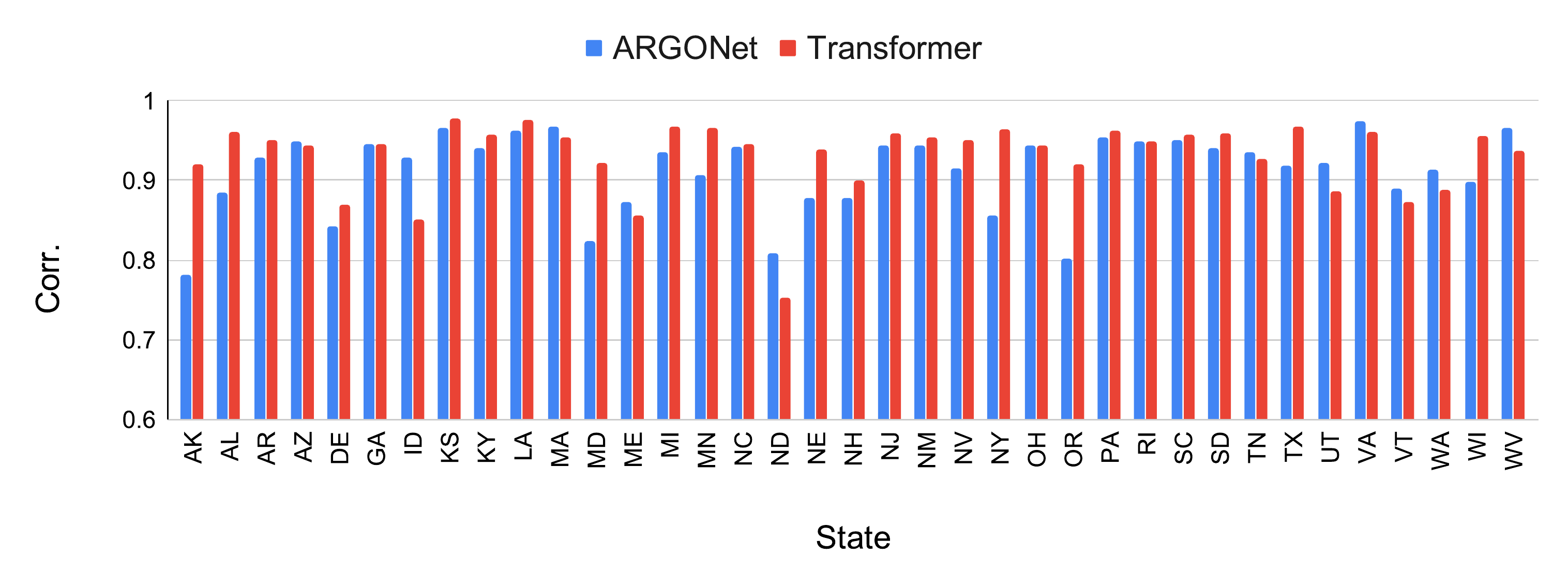}
\caption{Pearson Correlation of ARGONet and transformer models.}
\label{fig:argo_transformer_corr}
\end{figure*}

\begin{figure*}
\includegraphics[width=\textwidth]{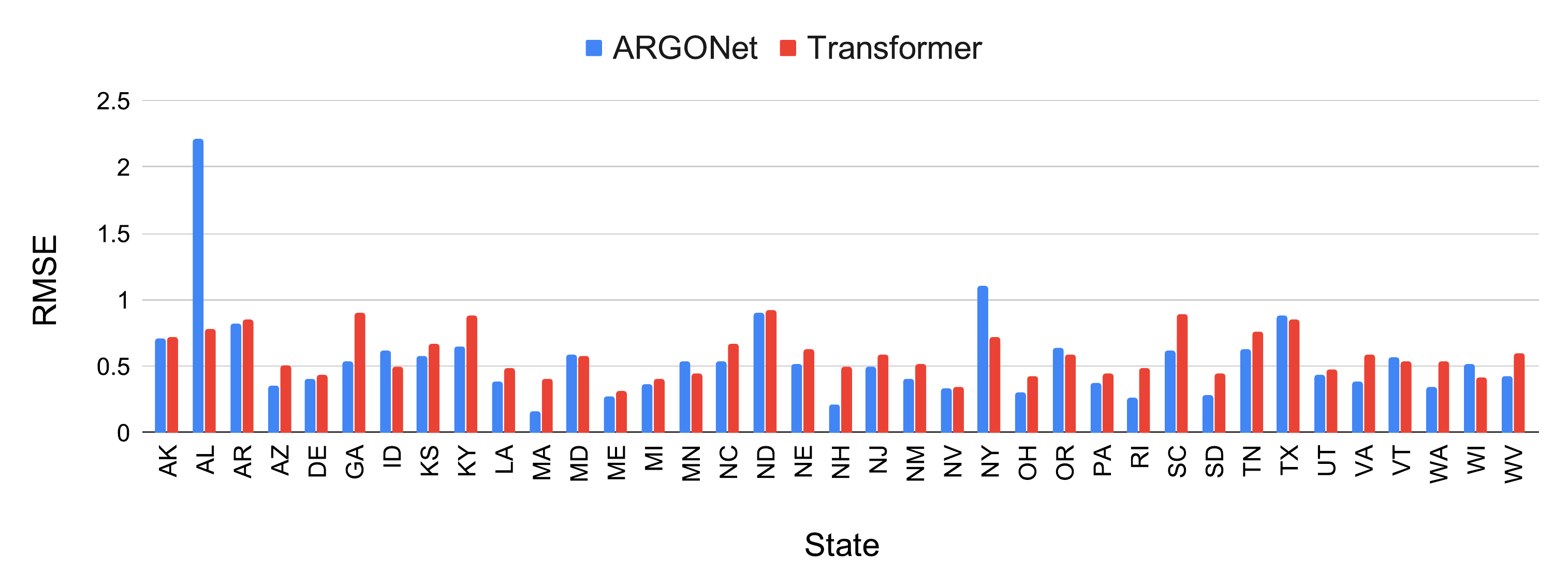}
\caption{RMSE of ARGONet and transformer models.}
\label{fig:argo_transformer_rmse}
\end{figure*}
In our first experiment, we tested whether
our Transformer-based model could predict the ILI ratio one-week ahead from 10 weeks of historical datapoints.
For evaluation, the trained global model performs one-step ahead prediction for each state using the testing data set.
Pearson correlation and root-mean-square error (RMSE) values were calculated for each state.

We compared the Transformer's performance with ARIMA, LSTM, and Seq2Seq with attention models.
Table ~\ref{table:exp_compare} summarizes the correlation coefficients and RMSEs for each method, as well as relative performance gain with respect to ARIMA method.
The comparison suggests that deep learning models overall outperform ARIMA for both correlation and RMSE.
Within the three deep learning approaches, the correlation coefficients are very similar with the Transformer-based model being slightly higher than LSTM and Seq2Seq with attention models.
In terms of RMSE, the Transformer model outperforms both LSTM and Seq2Seq with attention models, with relative RMSE decrease of 27 \% and 8.4 \%, respectively.
This analysis suggests that attention mechanisms contribute to forecasting performance, as Seq2Seq with attention and Transformer models outperform the plain LSTM model.
Additionally, the Transformer shows better forecasting performance compared to Seq2Seq with attention model,
suggesting that Transformer's self-attention mechanism can better capture complex dynamical patterns in the data compared to the linear attention mechanism used in Seq2Seq.
Interestingly, it's worth noting that Transformer exhibits the best metrics for US-level ILI forecasting (Pearson correlation = 0.984 and RMSE = 0.3318).
Since a single model is trained using data from all the states, it suggests that the model indeed can generalize various state-level patterns for country-level prediction.
\begin{table}[h!]
\centering
   \caption{Summary of model performances with relative change with respect to baseline model. }
   \label{table:exp_compare}
   \begin{tabular}{| c || c | c ||}
   \hline
   Model & Pearson Correlation & RMSE \\ [0.5ex]
   \hline\hline
   ARIMA &
     \begin{tabular}{@{}c@{}}0.769 \\ \small{(+0 \%)} \end{tabular} &
     \begin{tabular}{@{}c@{}} 1.020 \\ \small{(-0 \%)} \end{tabular}\\
     \hline
   LSTM &
     \begin{tabular}{@{}c@{}}0.924 \\ \small{(+19.9 \%)} \end{tabular} &
     \begin{tabular}{@{}c@{}} 0.807 \\ \small{(-20.9 \%)} \end{tabular}\\
     \hline
   Seq2Seq+attn &
     \begin{tabular}{@{}c@{}}0.920 \\ \small{(+19.5 \%)} \end{tabular} &
     \begin{tabular}{@{}c@{}} 0.642 \\ \small{(-37.1 \%)} \end{tabular}\\
     \hline
   Transformer &
     \begin{tabular}{@{}c@{}}0.928 \\ \small{(+20.7 \%)} \end{tabular} &
     \begin{tabular}{@{}c@{}} 0.588 \\ \small{(-42.4 \%)} \end{tabular} \\
     \hline
   \end{tabular}
\end{table}
\subsection{One-step-ahead Forecasting Using Feature Vectors}
We next tested whether our Transformer-based model can learn from multiple features (i.e., multivariate time series data) for ILI forecasting.
In the United States, the flu season usually starts in early October and peaks between January and Februray.
We hypothesized that week number is an informative signal for the model.
Therefore we introduced "week number" as a time-indexed feature to the model.
Additionally, we included the first and second order differences of the time series as two explicit numerical features in the model.

Our results suggest that including these features improves model performance (mean Pearson correlation:  0.931, mean RMSE = 0.585).
However, the improvement is not significant compared to the Transformer model using ILI data alone.
This suggests that the additional features are likely to encode little new information to the model.
That is, the introduced first and second order difference features are likely to be redundant
if the Transformer-based model is able to rely on the self-attention mechanism to learn short and long-range dependencies from the ILI time series.

We compared our results with the ILI forecasting data by ARGONet \cite{Lu2019}, a state-of-the-art ILI forecasting model in the literature.
Figure ~\ref{fig:argo_transformer_corr} and figure ~\ref{fig:argo_transformer_rmse} show the correlation and RMSE values of ARGONet and our transformer results.
Overall, the Transformer-based model performs equally with ARGONet, with the mean correlation slightly improved (ARGONet: 0.912, Transformer:  0.931), and mean RMSE value slightly degraded (ARGONet: 0.550, Transformer: 0.593).

\subsection{Forecasting Using Time Delay Embedding}
In this section, we tested whether the Transformer-based model can directly model phase space of a dynamical system.
To that end, we constructed time delay embeddings (TDEs) from historical ILI data, since
TDEs (with sufficient dimensionality) are topologically equivalent to the unknown phase space of dynamical systems.
In other words, compared to ILI data, which are observed scalar variables, TDEs encode additional geometrical and topological information of the systems that governs the process of influenza infection and spreading.
Therefore, using TDEs should provide richer information compared to scalar time series input.

To verify this hypothesis, we constructed time delay embeddings of dimension 2 to 32 from ILI data and applied transformer-based ILI forecasting using TDEs as features.
Table ~\ref{tde:dim} summarizes the forecasting metrics with different TDE dimensions $d$. In all the experiments, we use $\tau = 1$ to construct TDEs.
Varying TDE dimensionality does not significantly alter Pearson correlation coefficients. The RMSE value reached minimum with dimensionality of 8.
This value is close to the optimal TDE dimensionality of 5 and 5-7 for forecasting chickenpox and measles \cite{Sugihara1990}.

\begin{table}[h!]
\centering
 \caption{Performance of Time Delay Embeddings}
 \label{tde:dim}
 \begin{tabular}{| c || c | c ||}
 \hline
 Dimension & Pearson Correlation & RMSE\\ [0.5ex]
 \hline\hline
 2 & 0.926 & 0.745 \\ [1ex]
 \hline
 4 & 0.929 & 0.778 \\ [1ex]
 \hline
 6 & 0.927 & 0.618 \\ [1ex]
 \hline
 8 & 0.926  & 0.605  \\ [1ex]
 \hline
 16 & 0.925 & 0.623 \\ [1ex]
 \hline
 32 & 0.925 & 0.804 \\ [1ex]
 \hline
 \end{tabular}
\end{table}